\documentclass{article}

\PassOptionsToPackage{numbers, compress}{natbib}

\usepackage[preprint]{neurips_2021}

\usepackage[utf8]{inputenc} 
\usepackage[T1]{fontenc}    
\usepackage[hyphens,spaces,obeyspaces]{url}
\usepackage{booktabs}       
\usepackage{amsfonts}       
\usepackage{nicefrac}       
\usepackage{microtype}      

\usepackage[table]{xcolor}         

\usepackage{graphicx}

\usepackage{comment}

\usepackage{amsmath}
\usepackage{multicol}
\usepackage{multirow}
\usepackage{svg}
\usepackage{supertabular}
\usepackage{wrapfig}

\usepackage{floatrow}
\usepackage{tabularx}

\usepackage{array}
\newcolumntype{P}[1]{>{\centering\arraybackslash}p{#1}}

\usepackage{mathtools}
\graphicspath{ {./} }

\usepackage{amssymb}

\DeclareMathOperator{\softmax}{softmax}

\title{Distilling Knowledge via Intermediate Classifiers}

\author{%
  Aryan Asadian\\
  Ontario Tech University\\
  Ontario, Canada \\
  \texttt{aryan.asadian@ontariotechu.net} \\
   \And
   Amirali Salehi-Abari \\
  Ontario Tech University \\
  Ontario, Canada \\
   \texttt{abari@ontariotechu.ca} \\
}

\begin{document}


\DeclarePairedDelimiterX{\infdivx}[2]{(}{)}{
  #1\;\delimsize|\delimsize|\;#2
}
\newcommand{\kld}{D_\mathit{KL}\infdivx}
\newcommand{\h}{\mathcal{H}\infdivx}
\DeclarePairedDelimiter{\norm}{\lVert}{\rVert}
\newcommand{\lkd}{\mathcal{L}_{\scriptscriptstyle \mathtt{KD}}}
\newcommand{\lh}{\mathcal{L}_{\scriptscriptstyle \mathtt{CE}}}
\newcommand{\lst}{\mathcal{L}_{\scriptscriptstyle \mathtt{S}}}
\newcommand{\ldih}{\mathcal{L}_{\scriptscriptstyle \mathtt{DIH}}}
\newcommand{\yt}{y_{\scriptscriptstyle \mathtt{true}}}
\newcommand{\bftheta}{\pmb{\theta}}
\newcommand{\bfx}{\mathbf{x}}
\newcommand{\calT}{\mathcal{T}}
\newcommand{\bfpsi}{\pmb{\psi}}
\newcommand{\bfphi}{\pmb{\phi}}
\newcommand{\pt}{\mathrm{P}^{\tau}_\mathrm{T}}
\newcommand{\ph}{\mathrm{P}^{\tau}_\mathrm{h}}
\newcommand{\ps}{\mathrm{P}^{\tau}_\mathrm{S}}
\newcommand{\ind}[1]{\mathds{1}\left[#1\right]}
\definecolor{lgreen}{rgb}{0.1, 0.55, 0.1}
\newcommand{\upa}{\textcolor{lgreen}{\uparrow}}
\newcommand{\downa}{\textcolor{red}{\downarrow}}

\maketitle

\begin{abstract}
  The crux of knowledge distillation 
  is to effectively train a resource-limited student model with the guide of a pre-trained larger teacher model. However, when there is a large difference between the model complexities of teacher and student (i.e., \emph{capacity gap}), knowledge distillation loses its strength in transferring knowledge from the teacher to the student, thus training a weaker student. To mitigate the impact of the capacity gap, we introduce \emph{knowledge distillation via intermediate heads}. By extending the intermediate layers of the teacher (at various depths) with classifier heads, we cheaply acquire a cohort of heterogeneous pre-trained teachers. The intermediate classifier heads can all together be efficiently learned while freezing the backbone of the pre-trained teacher.  The cohort of teachers (including the original teacher) co-teach the student simultaneously. Our experiments on various teacher-student pairs and datasets have demonstrated that the proposed approach outperforms the canonical knowledge distillation approach and its extensions.
\end{abstract}

\section{Introduction}\label{introduction}
Deep neural networks have exhibited state-of-the-art performance in various domains such as computer vision \citep{vision_cite} and natural language processing \citep{devlin2018bert}. However, these models notoriously contain many parameters, requiring large storage spaces and intensive computation resources for training and inference. These requirements have impeded the deployment of deep neural networks in resource-limited environments (e.g., mobile and embedded devices). To acquire more compact yet effective models, a broad range of solutions have been developed such as network pruning \citep{blalock2020state}, network quantization \citep{yang2019quantization},  design of efficient architectures \citep{tan2019efficientnet}, and knowledge distillation \citep{hinton2015distilling}.
In \emph{knowledge distillation}, a small student learns (and gains richer information) by mimicking pre-trained teacher's output probabilities \citep{hinton2015distilling} and/or its intermediate representations \citep{romero2014fitnets,zagoruyko2016paying}. However, the student's accuracy might downgrade when the teacher's and the student's representation capacities are extremely disparate (i.e., large \emph{capacity gap} between student and teacher) \citep{mirzadeh2020improved, residual_kd}.

Various solutions have been proposed to mitigate the capacity gap in knowledge distillation. Knowledge distillation with teacher assistants (TAKD) \citep{mirzadeh2020improved}  distills the knowledge from teacher to the student with the assistance of a sequence of teacher assistants. Hint distillation \citep{romero2014fitnets} transfers the teacher's intermediate knowledge to the student in addition to class probabilities. Although these approaches improve the conventional knowledge distillation accuracy, they suffer from some technical challenges (e.g., large intermediate dimensionality mismatch between teacher and student in hint distillation) and are computationally expensive (e.g., training a sequence of teacher assistants in TAKD).

Intermediate classifier heads have shown to be effective medium for transferring knowledge between intermediate representations \cite{tofd,patient,mhkd,dcm,grafting}. Motivated by these developments,  we propose a novel variant of knowledge distillation called \emph{distillation via intermediate heads (DIH)} to address the capacity gap challenges. Our approach extends the intermediate layers of the pre-trained teacher (at multiple depths) with lightweight classifier heads. Each intermediate classifier facilitates the representation of its mounted intermediate layer in a common language of class probabilities. These classifier heads are cheaply trained as they all share the backbone of the pre-trained and ``frozen'' teacher. Acting as a cohort of teachers, the classifier heads and the original teacher co-teach the student. Our experiments on various teacher-student pairs of models have shown that DIH outperforms the canonical knowledge distillation approach and its state-of-the-art extensions on various benchmarks. Our experiments also suggest that DIH can be used as a general training framework to improve testing accuracy, even when the teacher and student are identical.

\section{Related Work}
\label{related_work}
There has been growing interest in model compression techniques for compressing the larger neural networks to smaller ones. 
Model pruning, as an optimization framework, intends to reduce the unnecessary structure and parameters of a large model to gain an efficient smaller network \citep{blalock2020state}. 
Model quantization compresses the neural network by reducing the precision of its parameters at the expense of accuracy  \citep{yang2019quantization}. Designing efficient deep neural architectures \citep{tan2019efficientnet} intends to satisfy the specified accuracy requirement given  resource limitations. Knowledge distillation \citep{hinton2015distilling}, that can be combined with others, aims to preserve the accuracy of a large trained teacher model in a smaller student model.

\vskip 1.5mm
\noindent \textbf{Knowledge Distillation.} 
Knowledge distillation is a training framework in which a huge pre-trained teacher model transfers its knowledge to a small student network  \citep{bucilua2006model, hinton2015distilling} to preserve teacher's accuracy in a resource-limited student model. The student trained under this framework is expected to have higher accuracy compared to if it was trained by the conventional training framework. The student owes this success to the teacher's output class probabilities for each input data. The relative class probabilities tell the student how the teacher generalizes and views the similarities between classes for an input sample. Many different extensions to knowledge distillation are proposed to leverage intermediate representations \citep{romero2014fitnets,zagoruyko2016paying,patient}, shared representations \citep{YuLMD17,gift}, representation grafting \cite{grafting}, adversarial learning \citep{improved, NEURIPS2018_019d385e,wang-c}, and contrastive learning \citep{contrastive}. Some other variants have improved the robustness of the student by injecting noise in the teacher's outputs \citep{sau}, aggregating multiple teachers' knowledge by voting mechanisms \citep{multiple_ta}, deploying teacher assistants \citep{mirzadeh2020improved}, or allowing peers to teach each other \citep{zhang2018deep}.
Our work is closely related to those extensions of knowledge distillations, which exploit the knowledge of intermediate representations (e.g., \citep{romero2014fitnets,zagoruyko2016paying,patient,YuLMD17}). We attempt to address their practical challenges when (1) the dimension mismatch between student's and teacher's intermediate layers is large, (2) the choices of intermediate/shared layers are not obvious for the teacher and student models.  
\vskip 1.5mm
\noindent \textbf{Knowledge Distillation via Intermediate Classifiers.}  Intermediate classifier heads have recently shown to be an effective medium for transferring knowledge from teacher's intermediate representations to the student \cite{tofd,patient,mhkd,dcm,grafting}.\footnote{The concept of intermediate/auxiliary classifier heads has also been explored in Inception \citep{szegedy2015going} as regularizers, for understanding intermediate representations \citep{inter_probs} as \emph{probes}, and for reducing the inference time \citep{teerapittayanon2016branchynet}.} The current distillation approaches based on intermediate heads can fall into two categories. In \emph{one-to-one} approach, the teacher's knowledge at an intermediate layer is transferred to a corresponding intermediate layer of the student through two classifier heads: one for the teacher and the other one for the student (e.g., \citep{tofd,patient,mhkd}). The \emph{many-to-many} approach, mainly explored in online distillation, mounts multiple intermediate heads on each peer, and then each head distills knowledge from every other peers' heads \cite{dcm}. Both one-to-one and many-to-many approaches pose a few computational and technical challenges: (i) the choice for the placing student's intermediate heads is not straightforward unless the student and teacher share the similar architecture; (ii) the intermediate heads are complex with many parameters (e.g., containing convolutions, batch normalization, pooling layer, and a fully connected layer in \cite{tofd}), thus requiring retraining the teacher from scratch \cite{tofd}; and (iii) their hyperparameter tuning is difficult as more hyperparameters are involved in controlling the trade-off between losses of the student's and teacher's heads. These challenges make them underperform the canonical knowledge distillation, especially when the capacity gap is large between teacher and student (as shown in our experiments below). Our work attempts to address these challenges.

\vskip 1.5mm
\noindent \textbf{Our Work.} As installing intermediate heads on students are the source of the challenges i--iii (discussed above), we propose a novel \emph{many-to-one} approach for distillation via intermediate heads: By mounting multiple intermediate heads just on the teacher, the student distills the knowledge from the teacher and its intermediate layers in the common language of class probabilities. The teacher's intermediate heads are lightweight and can be cheaply fine-tuned by recycling the pre-trained teacher (i.e., without retraining the teacher). Our work could loosely be viewed as a type of curriculum learning \citep{bengio2009curriculum}. Instead of ordering the input data based on their difficulty, we simultaneously provide both easy and complicated representations via multiple intermediate classifier heads to the student. These representations with different levels of complexity allow the student to learn from the teacher based on its representation capacity.
\section{Distillation via Intermediate Heads}
\label{dih}
In this section, we first review the necessary background for the canonical teacher-student setting \citep{hinton2015distilling} of knowledge distillation. Then, we detail how our proposed solution extends this setting.
\vskip 1.5mm \label{review_kd}
\noindent \textbf{Review of Knowledge Distillation.} The crux of knowledge distillation is to train a small \emph{student} network from the output of a larger pre-trained \emph{teacher} network \citep{hinton2015distilling} (in addition to true labels).  When the teacher assigns high probability to multiple classes for a given input (e.g., image), this information suggests that the input shares some common patterns for those classes. By matching its outputs with those of the teacher, the student learns finer patterns learned by the teacher.  

Consider the classification task over $C$ classes, given a dataset $\mathcal{D}$ with input-label pairs $(\mathbf{x}, y)$ and $y \in \{1,\dots, C\}.$ Assume that the teacher network $\mathrm{T}$ is trained on the same data and its output for an input $\mathbf{x}$ is given by
$
\pt(\mathbf{x}) = \softmax\left(\frac{\mathbf{z}_{\scriptscriptstyle T}}{\tau}\right),
$
where $\mathbf{z}_{\scriptscriptstyle T}$ is the activation vector before the softmax layer, and $\tau \geq 1$ is the temperature parameter. This parameter intends to soften the output probabilities to further preserve the information learned by the teacher (e.g., the relationship of classes). The higher $\tau$ is, the more output probabilities are softened.\footnote{The teacher is usually trained with $\tau=1$, whereas its $\tau$ is set to a higher value during the distillation. The higher value allows the teacher to focus more on conveying the inter-class similarities to the student rather than absolute values of probabilities, especially when it has very high confidence in its output.} Similarly, the student network $S$ outputs its soften probabilities over classes by
$
\ps(\mathbf{x}) = \softmax\left(\frac{\mathbf{z}_{\scriptscriptstyle S}}{\tau}\right).
$ The student $S$ can learn from the teacher $T$ by minimizing the \emph{knowledge distillation loss}:
\begin{equation}
    \lkd(S | T, \bfx) = \tau^2 \h[\Big]{\pt(\bfx)}{\ps(\bfx)},
    \label{eq:loss_kd}
\end{equation}
where $\h{\mathbf{p}}{\mathbf{q}} = -\sum_i p_i \log q_i$ is the cross-entropy of  the \emph{approximated distribution} $\mathbf{q}$ relative to the \emph{target distribution} $\mathbf{p}$. Here, cross-entropy measures the extent the student S's probability distribution $\ps(\bfx)$ differentiates from the target teacher T's distribution $\pt(\bfx)$ for input $\bfx$.\footnote{The knowledge distillation loss in Eq.~\ref{eq:loss_kd} can be equivalently defined by replacing the cross-enropy $\h{\mathbf{p}}{\mathbf{q}}$ with the Kullback-Leibler (KL) divergence distance $\kld{\mathbf{p}}{\mathbf{q}}=\sum_i p_i \log \frac{p_i}{q_i}$. This equivalence is because $\h{\pt(\bfx)}{\ps(\bfx)} = \kld{\pt(\bfx)}{\ps(\bfx)} + H (\pt(\bfx)) $ and the entropy  $H (\pt(\bfx))$ is a constant.} The scaling $\tau^2$ in this loss allows us to keep its gradient magnitudes approximately constant when the temperature $\tau$ changes \citep{hinton2015distilling}.

Given input $\bfx$, the student $S$ can learn from both its true label $\yt$ (i.e., hard target) and the teacher $T$'s output for input $\bfx$ (i.e., soft target) by minimizing the \emph{student loss}:
\begin{equation}
\lst (S | T, \bfx) = \alpha \lkd(S | T, \bfx)  + (1-\alpha) \lh(S|\bfx, \yt),
\label{eq:student}    
\end{equation}
where $\lh(S|\bfx, \yt)$ is a typical cross-entropy loss between student S's probability output $\ps(\bfx)$ and the degenerate data distribution for the true label $\yt$.\footnote{For the student $S$'s output probability $\ps(\bfx)$ in the cross-entropy loss $\lh$, we always set $\tau =1$ as with the original Knowledge distillation approach \citep{hinton2015distilling}.} Here, the hyperparameter $\alpha$ controls the trade-off between the two losses (i.e., the balance between hard and soft targets).

\begin{wrapfigure}{r}{0.53\textwidth}
\begin{center}
    \includegraphics[width=1\textwidth]{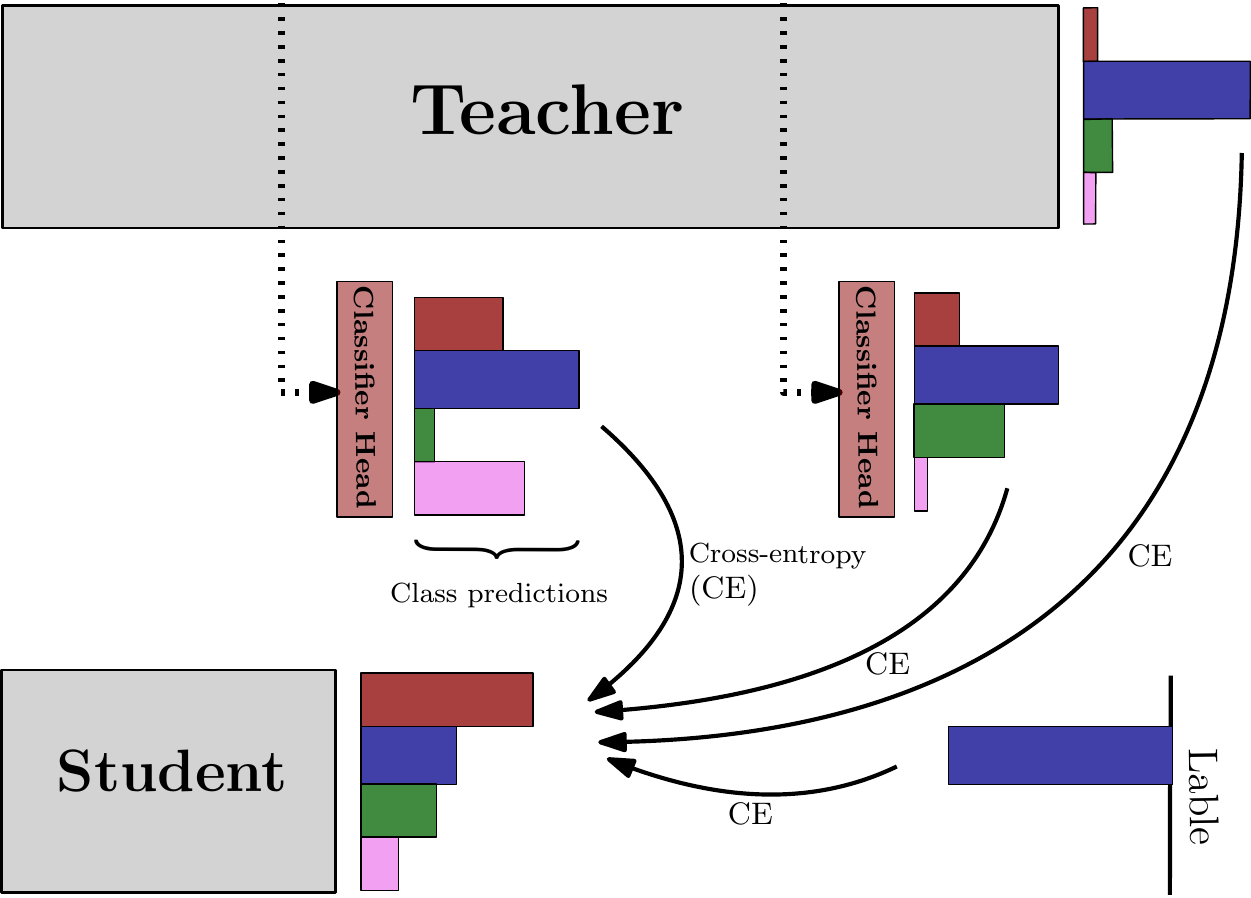}
        \caption{Distillation via Intermediate Heads (DIH). The teacher is equipped with intermediate classifier heads at various depths. These new heads are trained while the backbone is frozen. A cohort of teachers, including all the intermediate classifiers and the original teacher, distills knowledge to the student.}
       \label{fig:dih_scheme}
\end{center}
\end{wrapfigure}

\vskip 1.5mm
\noindent \textbf{Distillation via Intermediate Heads (DIH).} 
We propose \emph{Distillation via Intermediate heads (DIH)} to leverage intermediate representations of the teacher in knowledge distillation and efficiently address the capacity gap issue. Fig.~\ref{fig:dih_scheme} illustrates the general scheme of our proposed framework. The central idea behind DIH is to cheaply acquire a cohort of teachers with various model complexities by recycling the pre-trained teacher. To create such a cohort, DIH extends the intermediate layers of the teacher (at various depths) with classifier heads. Then, it efficiently trains these intermediate classifier heads all together while freezing the backbone of the original teacher. Then, the student distills the knowledge simultaneously from the cohort of teachers (including the original teacher).

The DIH approach selects $k$ \emph{mounting positions} from the teacher $T$ to mount $k$ intermediate classifier heads. These mounting positions can be any arbitrary intermediate layer or the output of a submodule (e.g., a stack of multiple convolutions, pooling, dropout, and batch normalization layers).  The output of the intermediate classifier head $h$ at the mounting position $i$ is given by
\begin{align}
    &\ph(\mathbf{x}) = \softmax\left(\frac{\mathbf{z}_h(\mathbf{x})}{\tau}\right),
\end{align}
and $\mathbf{z}_h(\mathbf{x}) = g\left(\mathbf{W}^{(h)}\mathbf{a}_i(\mathbf{x}) + \mathbf{b}^{(h)}\right)$, where $\mathbf{a}_i(\mathbf{x})$ is the activation vector of the mounting position $i$ of teacher $T$ for the input $\mathbf{x}$. Here, $\mathbf{W}^{(h)}$, $\mathbf{b}^{(h)}$, and $g$ are the linear transformation weight matrix, bias vector, and non-linear activation function (e.g., Relu) for the classifier head  $h$, respectively. The classifier head $h$ converts the knowledge of its mounting position (i.e., intermediate layer of the teacher) to class predictions. We learn classifier heads using a cross-entropy loss, when the teacher's backbone is frozen. The training of classifier heads are computationally cheap for two reasons: (i) each classifier head $h$ has relatively few parameters compared to the large size of teacher;\footnote{For the head $h$ mounted at intermediate layer $i$, the number of parameters are $(N_i+1) \times C$, where $N_i$ is the size of the activation vector at mounting position $i$ and $C$ is the number of classes.} and (ii) the computations within teacher's backbone can be reused for training all $k$ heads at the same time.      

The original pre-trained teacher $T$ along with the $k$ trained classifiers heads form the cohort of $k+1$ teachers $\calT$. This cohort of teachers offers few practical and computational advantages for knowledge distillation. It encompasses heterogeneous teachers in terms of their model complexities.\footnote{
The higher intermediate head results in the teacher with the higher model complexity.} The inference time of this cohort of teachers is comparable to that of the original teacher $T$ as all intermediate heads share the backbone of the original teacher. The student $S$ distills knowledge from the cohort $\calT$ by minimizing 
\begin{equation}
    \ldih (S | \calT, \bfx) = \frac{1}{k+1} \sum_{M \in \calT} \lkd (S | M, \bfx),
    \label{eg:loss_dih}
\end{equation}
where the knowledge distillation loss $\lkd (S | M, \bfx)$ is computed by Eq.~\ref{eq:loss_kd}. This loss is the average of distillation losses for student $S$ distilling knowledge from cohort $\calT$ (including the original teacher and $k$ intermediate heads). To benefit from hard targets (i.e., true labels), the student $S$ minimize student loss for input $\bfx$: 
\begin{equation}
\lst (S | \calT, \bfx) = \alpha \ldih(S | \calT, \bfx)  + (1-\alpha) \lh(S|\bfx, \yt),
\label{eq:student-DIH}    
\end{equation}

where $\lh(S|\bfx, \yt)$ is a typical cross-entropy loss between student S's probability output and the true label $\yt$, and $\alpha$ controls the trade-off between the two losses.

\section{Empirical Experiments} \label{experiments_section}
We evaluate the distillation via intermediate heads (DIH) framework by comparing it against canonical knowledge distillation and its state-of-the-art extensions, which leverage intermediate representations.
\subsection{Experimental Setup} \label{exp_setup}
\vskip 1.5mm
\noindent \textbf{Datasets.} We use three standard datasets of CIFAR-10, CIFAR-100 \citep{krizhevsky2009learning}, and Tiny-ImageNet~\citep{tiny_imagenet}.\footnote{The ImageNet dataset was not accessible for our research.} CIFAR-10 and CIFAR-100 datasets include 32x32 RGB images for $10$ and $100$ classes, respectively. Each dataset had $50K$ training and $10K$ testing images. Tiny-ImageNet has 110K 64x64-colored samples for $200$ classes, with the split of $100K$ and $10K$ for training and testing. All training and testing datasets are balanced (i.e., the number of images per class is the same within the dataset). For all datasets, similar to \citep{BMVC2016_87,zagoruyko2016paying,mirzadeh2020improved}, the images are augmented by the combination of horizontal flips, 4 pixels padding, and random crops. We normalized the images by their mean and standard deviation. 
\begin{table*}[tb]
\begin{center}
\caption{Experimented networks with their numbers of parameters (for CIFAR-100) and mounted heads k. The models are ordered from smallest to largest, left to right.}
\label{tab:model_sizes}
\begin{tabular}{p{1cm}p{1cm}p{1cm}p{1.4cm}p{1.23cm}p{1cm}p{1cm}p{1cm}}
\toprule
{Res8}&{Res14}&{Res20}&{WR28-2}&{Res110}&{VGG11}&{Res18}&{Res34}\\
\midrule
0.08M&0.18M&0.28M&1.48M&1.74M& 9.27M& 11.22M& 21.32M\\
k=3&k=3&k=3&k=3&k=3&k=5&k=4&k=4\\
\bottomrule
\end{tabular}
\end{center}
\end{table*}

\vskip 1mm
\noindent \textbf{Networks.} We use three classes of architectures with different depths as teacher model: ResNet family \citep{He_2016_CVPR}, VGG11 \citep{simonyan2014very}, and Wide Residual Networks~\citep{wide_resnet}. Our choice is motivated by the prevalence of these models in computer vision tasks. 
We deploy the ResNet family with varying numbers of stacked residual blocks to control model complexities (see Table \ref{tab:model_sizes}). We also used VGG11, equipped with batch normalization, as a straightforward deep teacher model. For ResNet family teachers, we mounted the classifier heads  after each group of same-dimensional residual blocks (e.g., ResNet-34 has four heads). In the VGG teacher, we added each head after any max-pooling layer. Table \ref{tab:model_sizes} shows the size of models (i.e., the number of parameters) and also the number of added classifier heads $k$.

\begin{table}[tb]
\caption{Test accuracy (\%) of Res8 student network on various teachers and datasets. The student is trained by DIH (ours), canonical KD, or regular cross-entropy (CE). Imp. stands for improvement between the best (in bold) and the second best (in italics). Average over three runs.}
    \begin{center}
    \begin{tabular}{lP{0.607cm}P{0.607cm}P{0.607cm}P{0.607cm}P{0.607cm}P{0.607cm}P{0.607cm}P{0.607cm}P{0.607cm}P{0.607cm}P{0.607cm}P{0.607cm}}
    \toprule
    &
    \multicolumn{4}{c}{CIFAR-10}&
   \multicolumn{4}{c}{CIFAR-100}
   &\multicolumn{4}{c}{Tiny-ImageNet }\\ 
   \cmidrule[0.7pt](lr){2-5} 
   \cmidrule[0.7pt](lr){6-9} 
   \cmidrule[0.7pt](lr){10-13}
   Teacher&CE&KD&DIH&Imp.&CE&KD&DIH&Imp.&CE&KD&DIH&Imp.\\
\cmidrule[0.7pt](lr){1-1} 
\cmidrule[0.7pt](lr){2-5} 
   \cmidrule[0.7pt](lr){6-9} 
   \cmidrule[0.7pt](lr){10-13}
   WR28-2
   &88.19&\textit{88.82}&\textbf{89.89}&1.07&60.47&\textit{60.78}&\textbf{63.32}&2.54&40.45&\textit{40.70}&\textbf{43.89}&3.19\\
Res110
&88.19&\textit{89.30}&\textbf{89.44}&0.14&60.47&\textit{62.31}&\textbf{63.36}&1.05&40.45&\textit{40.47}&\textbf{42.25}&1.78\\
VGG11
&88.19&\textit{88.41}&\textbf{89.91}&1.50&60.47&\textit{61.10}&\textbf{63.79}&2.69&40.45&\textit{40.76}&\textbf{43.78}&3.02
                   \\
Res34
&88.19&\textit{89.26}&\textbf{90.00}&0.74&60.47&\textit{61.68}&\textbf{63.06}&1.38&\textit{40.45}&40.01&\textbf{43.00}&2.55\\
\bottomrule
\end{tabular}
\label{tab:cifars_becnhmark}
\end{center}
\end{table}
\definecolor{gold}{rgb}{0.83, 0.69, 0.22}
\definecolor{silver}{rgb}{0.79, 0.75, 0.73}
\definecolor{bronze}{rgb}{0.8, 0.5, 0.2}
\def\1#1&#2&#3&#4&#5&#6&#7&#8\\{#7 &#6 &#5 &#4 &#3 &#2 &#1 \\}
\begin{table*}[t]
\begin{center}
\caption{Test accuracy (\%) of various student-teacher pairs on CIFAR-100 datasets.
The best, second, and third are shown with gold, silver, and bronze backgrounds, respectively.
$\upa$ and $\downa$ denote better and worse than KD. Capacity Ratio is the ratio of the number of parameters in the teacher model to the number of the student's parameters  DIH (ours) is the only method that consistently has outperformed KD for all teacher-student pairs with any capacity ratio. Average over three runs.}
 \label{tab:kd_compare} 
\begin{tabular}{lllllllll}
\toprule
Teacher&\1Res34&VGG11&Res34&Res110&WR28-2&Res34&Res20&Res110\\
Student&\1Res8&Res8&Res20&Res8&Res8&WR28-2&Res8&WR28-2\\
Capacity Ratio&\1 266.50&115.80&76.14&21.75&18.50&15.86&3.50&1.17\\
\midrule
CE&\160.47~~~&60.47~~~&69.30~~~&60.47~~~&60.47~~~&70.08~~~&60.47~~~&70.08~~~\\
KD~\cite{hinton2015distilling}&\161.68~~~&61.10~~~&69.10~~~&\cellcolor{silver}62.31~~~&60.78~~~&69.84~~~&61.22~~~&70.71~~~\\
FitNets~\cite{romero2014fitnets}&\1 61.53$\downa$&\cellcolor{silver}61.37$\upa$& 67.20$\downa$&\cellcolor{bronze}61.55$\downa$& 60.74$\downa$&70.81$\upa$&\cellcolor{silver}61.69$\upa$& 71.08$\upa$\\
TAKD~\cite{mirzadeh2020improved}&\1\cellcolor{bronze}61.73$\upa$& 60.99$\downa$& 69.22$\upa$& 61.41$\downa$&\cellcolor{bronze}61.05$\upa$& 68.47$\downa$& 61.52$\upa$& 70.64$\downa$\\
TOFD~\cite{tofd}&\1\cellcolor{silver}61.76$\upa$&\cellcolor{bronze}61.26$\upa$& 69.55$\upa$& 61.47$\downa$& 61.86$\upa$&\cellcolor{silver}72.14$\upa$& 61.44$\upa$&\cellcolor{silver}73.08$\upa$\\
AT~\cite{zagoruyko2016paying}&\1 60.36$\downa$&54.91$\downa$& 69.39$\upa$& 61.05$\downa$& 60.34$\downa$& 70.97$\upa$& 60.85$\downa$&\cellcolor{bronze}72.31$\upa$\\
MHKD~\cite{mhkd}&\1 60.66$\downa$& 60.54$\downa$&\cellcolor{gold}71.00$\upa$& 61.16$\downa$& 60.94$\upa$& 70.39$\upa$& 60.03$\downa$& 71.90$\upa$\\
CRD~\cite{crd}&\1 60.79$\downa$& 60.66$\downa$&\cellcolor{bronze}70.48$\upa$& 61.04$\downa$&\cellcolor{silver}61.60$\upa$&\cellcolor{gold}76.49$\upa$&\cellcolor{bronze}61.62$\upa$&\cellcolor{gold}76.90$\upa$\\
DIH (ours)&\1\cellcolor{gold}63.06$\upa$&\cellcolor{gold}63.79$\upa$&\cellcolor{silver}70.66$\upa$&\cellcolor{gold}63.36$\upa$&\cellcolor{gold}63.32$\upa$&\cellcolor{bronze}71.41$\upa$&\cellcolor{gold}63.11$\upa$& 71.78$\upa$\\
\bottomrule
\end{tabular}
\end{center}
\end{table*}
\vskip 1.5mm
\noindent \textbf{Training and hyperparameters.} We use PyTorch \citep{NEURIPS2019_9015} for implementation and experiments.
\footnote{The code is available at \url{https://github.com/aryanasadianuoit/Distilling-Knowledge-via-Intermediate-Classifiers}.}
The teacher and student models (for any knowledge distillation approach) are trained by SGD optimizer with Nesterov momentum of $0.9$ and weight decay of $5\times10^{-4}$. We train the models for $200$ epochs with the batch size of $128$ and the initial learning of $0.1$, multiplied by $0.2$ every $60$ epochs. The hyperparameters $\alpha$ and $\tau$ for KD and DIH are selected by grid search with multiple runs: $\alpha= 0.1$ and $\tau=5$ on all datasets for both DIH and KD with an exception of  $\tau=4$ for KD on CIFAR-100. 
\subsection{Results}
\begin{figure}\CenterFloatBoxes
\begin{floatrow}
\ffigbox[\FBwidth]
{%
  \caption{Output of Res110's heads on a CIFAR-10 image of plane.}
  \label{fig:r110}
}
{%
  \includegraphics[width=0.327\textwidth]{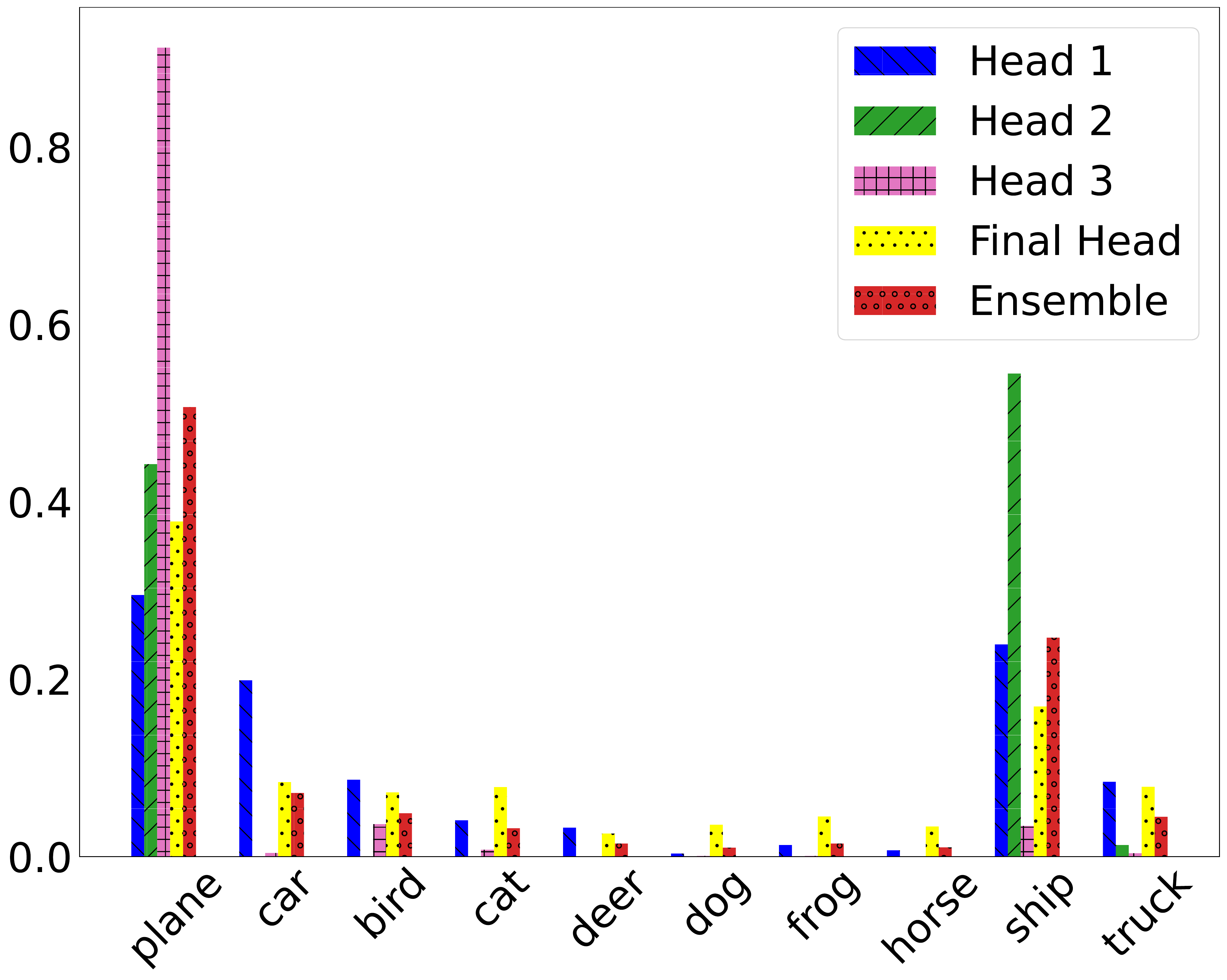}
 }
\killfloatstyle\ttabbox[\Xhsize]{%
\caption{Average entropy and KL Divergence of heads for Res110 teacher on CIFAR-10 training images, $\tau=5$.}
\label{tab:entropy-Kl} %
}{%
\begin{tabular}{lllll}
\toprule
&Head 1&Head 2&Head 3&Main\\
Entropy & 1.785 & 0.272 & 0.330 & 1.837\\
\midrule
Head 1&0&5.118&2.139&0.261\\
Head 2 &1.040&0&1.126&0.972\\
Head 3 &1.013&1.398&0&0.599\\
Main &0.282&5.492&1.422&0\\
\bottomrule
\end{tabular}
}
\end{floatrow}
\end{figure}

\noindent \textbf{Improvement on knowledge distillation.}
We first examine if distillation via intermediate heads (DIH) improves canonical knowledge distillation (KD) \citep{hinton2015distilling} over different sets of student-teacher pairs with a relatively large capacity gap. For this purpose, we trained Res8 as a small student with teachers of various depths and architectures. Table~\ref{tab:cifars_becnhmark} shows the accuracy of students for KD, DIH, and regular cross-entropy (CE) training (i.e., no distillation) for three datasets. DIH has improved both KD and CE in all datasets for any teacher-student pairs. The improvements are more noticeable for more complex datasets (e.g., Tiny-ImageNet and CIFAR-100) and when the capacity gap is larger. For example, compare the absolute improvements between Res110 and Res34 teachers: 0.14\% vs. 0.74\% in CIFAR-10 and 1.05\% vs. 1.38\% in CIFAR-100. We also note that KD fails to improve the student model when both the number of classes and the capacity gap are large. For example, KD optimizes the Res8 student poorly with Res34 teacher in Tiny-ImageNet when compared to regular cross-entropy ($40.01\%$ vs. $40.45\%$). However, DIH outperforms both CE and KD using the same settings by a large margin of $2.55$ and $2.99$, respectively.  
\vskip 1mm
\noindent \textbf{DIH vs. KD variants.}
We also study how DIH compares to other variants and extensions of knowledge distillation, which either take the advantage of intermediate representations for knowledge distillation or intend to address the capacity gap issue: \emph{FitNets} \citep{romero2014fitnets}, Attention distillation (\emph{AT}) \citep{zagoruyko2016paying}, Teacher Assistant Knowledge Distillation (\emph{TAKD}) \citep{mirzadeh2020improved}, Multi-head Knowledge Distillation (\emph{MHKD}) \citep{mhkd}, Task-Oriented Feature Distillation (\emph{TOFD}) \citep{tofd}, and Contrastive Representation Distillation \cite{crd}.\footnote{For FitNets, we trained the student for $60$ and $200$ epochs in the first and second stages, respectively.}
Note that MHKD and TOFD, similar to DIH, benefit from intermediate classifier heads, and TAKD is specifically designed to address the capacity gap. We experiment on various teacher-student pairs from different network families while varying the capacity gap. We measure the capacity gap by \emph{capacity ratio}, given by the ratio of the number of parameters in the teacher model to the number of the student's parameters. 
Table ~\ref{tab:kd_compare} shows the result of this set of experiments.  DIH outperforms other methods in all teacher-student pairs by a relatively large margin, with two exceptions. 
For two teacher-student pairs of (Res34, Res20) and (Res34, WR28-2), DIH ranks second and third among the counterparts. DIH is the only method that consistently has outperformed  KD for all teacher-student pairs with any capacity ratio. TOFD has also demonstrated consistent performance in outperforming KD for all teacher-student pairs except for the Res110-Res8 pair. We note that CRD has performed very well for relatively small capacity ratios (e.g., $< 21.75$), but its performance downgraded for large capacity ratios (e.g., $>100$). TAKD had mixed performance for both small and large capacity ratios and even for some teacher-student pairs has underperformed KD. FitNets had inconsistent performance by outperforming over KD for only three teacher-student pairs. The distillation of Res110 to Res8 seems to be the most challenging setting when none of the approaches (except DIH) could outperform the canonical knowledge distillation. 
As TAKD shares our goal in addressing the capacity gap, we run a set of experiments on TAKD with the Res34 teacher and the Res8 student. We vary the chain of teacher assistants for TAKD to optimize its performance. As shown in Table~\ref{tab:takds_comapre}, TAKD seems to be sensitive to the chain of teacher assistants. To outperform the KD, it requires training six teacher assistant models, thus making it computationally expensive. However, DIH firmly outperforms all combinations of TAKD with a considerable margin while requiring a small computational overhead (for fine-tuning the intermediate heads). 
\begin{table*}[tb]
\begin{center}
\caption{Test Accuracy (\%) for TAKD with various chains of teacher assistants, canonical KD, and DIH. The Res8 student distills from Res34 teacher on CIFAR-100 dataset. Average over three runs.}
 \label{tab:takds_comapre} 
\begin{tabular}{llc}
\toprule
Method &Teacher Assistants& Accuracy\\
\midrule
KD& - &61.68\\
\midrule
TAKD&Res18&60.82\\
TAKD&Res18$\rightarrow$Res110&60.73\\
TAKD&Res18$\rightarrow$Res110$\rightarrow$Res56&61.01\\
TAKD&Res18$\rightarrow$Res110$\rightarrow$Res56$\rightarrow$Res32&61.41\\
TAKD&Res18$\rightarrow$Res110$\rightarrow$Res56$\rightarrow$Res32$\rightarrow$Res20&61.60\\
TAKD&Res18$\rightarrow$Res110$\rightarrow$Res56$\rightarrow$Res32$\rightarrow$Res20$\rightarrow$Res14&61.73\\
\midrule
DIH&-&\textbf{63.06}\\
\bottomrule
\end{tabular}
\end{center}
\end{table*}
\vskip 1.5mm
\noindent \textbf{A deeper look into DIH.}
We aim to find out how powerful the ensemble of the intermediate classifier heads is when compared to DIH. We hope this investigation sheds light on how DIH works. We found that a student trained with an ensemble averaging of all classifier heads (including intermediate and the final head) is weaker than the same student that mimicked each classifier head separately. This observation is similar to what is reported for deep mutual learning \citep{zhang2018deep}: when a cohort of peers train each other, the distillation between each pair of peers is more powerful than the knowledge transfer between a peer and an ensemble of other peers. We believe that the ensemble averaging removes inter-class similarity information (see Fig.~\ref{fig:r110} as an example).  
Through our investigation, a few interesting observations arise which hold for all teachers regardless of their size and the architecture: (i)  both the teacher head and the ensemble averaging of all the heads have very similar and relatively smooth probability distribution; (ii) the intermediate classifier heads exhibit the various extent of confidence in their output probabilities (e.g., Head 1 and 3 with low and high confidence, resp., in Fig.~\ref{fig:r110}); and (iii) each head has relatively distinct probability distributions. While observation (i) explains why ensemble averaging is not better than canonical knowledge distillation in our settings, the observations (ii)--(iii) shed light on how DIH works. We further investigate these two observations from the information theory lenses.

\begin{wrapfigure}{r}{0.37\textwidth}
\vspace{-12pt}
\begin{center}
    \includegraphics[width=0.9\textwidth]{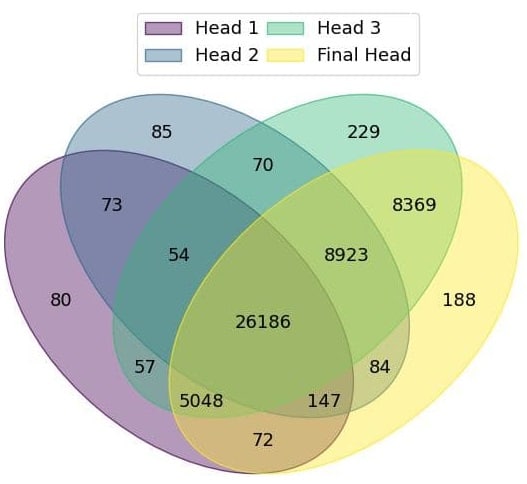}
        \caption{The number of correct predictions by each head in the Res110 teacher on CIFAR-10 training dataset.}
      \label{fig:overthinkdigram}
\end{center}
\end{wrapfigure}
The information-theoretic analyses show us that each head offers different knowledge for distillation; however, it is unclear if each head's knowledge is useful for distillation. Thus, we ask if each head can provide useful representations that are not provided by other heads. To answer this question, as heads output class probabilities, we study if a head can correctly classify some training instances while other heads fail to classify them correctly. If such instances exist for each head, the head's representation for those instances is unique and useful for knowledge distillation.
Figure ~\ref{fig:overthinkdigram} shows the Venn diagram of correct predictions for all heads on the CIFAR-10 training dataset with the Res110 teacher. We notice that each head has some instances that it exclusively could classify them correctly: $80$ for Head 1, $85$ for Head 2, $229$ for Head 3, and $188$ for Final Head. Surprisingly, the shallowest head (e.g., Head 1) can uniquely provide more ``correct'' representations for 80 instances. This observation highlight the value of each head (even the shallow ones) for knowledge distillation, while the deeper heads might be more prone to \emph{overthinking} \citep{overthinking}.

\vskip 1mm
\noindent \textbf{Ablation studies.}
To quantify the extent each head contributes to the success of DIH, we conduct a complete ablation study where in one extreme all heads are ``off'' (i.e., regular cross-entropy) and in another extreme, all heads are ``on.'' Table~\ref{tab:ablation_best_head} show the result of our ablation studies for training the Res8 student with the Res110 teacher. The addition of heads usually improves the distillation with a few exceptions (e.g., when H1 is added to the regular cross-entropy). The deeper heads can improve the distillation more than the shallower heads.  However, the best combination comprises all the classifier heads, confirming that DIH benefits from both shallow and deep heads.  

\begin{table}[tb]
\caption{Ablation study on DIH, CIFAR-100 dataset, Res110 teacher with four heads, Res8 student. Accuracy (\%) is the average of three runs. The best and  second best are in \textbf{bold} and \textit{italic}, respectively. The $\bullet$ and $\circ$ indicates ``on'' and ``off.''}
    
    \begin{center}
    \begin{tabular}{cccccccccc}
    \toprule
   $H_1$&$H_2$&$H_3$&$Main$&Accuracy& $H_1$&$H_2$&$H_3$&$Main$&Accuracy\\
   \cmidrule[0.7pt](lr){1-5} \cmidrule[0.7pt](lr){6-10}
   $\circ$&$\circ$&$\circ$&$\circ$&$60.47$&$\circ$&$\bullet$&$\bullet$&$\circ$&62.30\\
   $\bullet$&$\circ$&$\circ$&$\circ$&$59.60$&$\circ$&$\bullet$&$\circ$&$\bullet$&62.53\\
   $\circ$&$\bullet$&$\circ$&$\circ$&$60.61$&$\circ$&$\circ$&$\bullet$&$\bullet$&62.25\\
   $\circ$&$\circ$&$\bullet$&$\circ$&$60.49$&$\bullet$&$\bullet$&$\bullet$&$\circ$&62.54\\
   $\circ$&$\circ$&$\circ$&$\bullet$&$62.31$&$\bullet$&$\bullet$&$\circ$&$\bullet$&62.83\\
   $\bullet$&$\bullet$&$\circ$&$\circ$&$61.73$&$\bullet$&$\circ$&$\bullet$&$\bullet$&\textit{62.89}\\
   $\bullet$&$\circ$&$\bullet$&$\circ$&61.80&$\circ$&$\bullet$&$\bullet$&$\bullet$&62.79\\
   $\bullet$&$\circ$&$\circ$&$\bullet$&62.30&$\bullet$&$\bullet$&$\bullet$&$\bullet$&\textbf{63.36}\\
\bottomrule
\end{tabular}
\label{tab:ablation_best_head}
\end{center}
\vspace{-8pt}
\end{table}
\section{Conclusion} \label{conclusion_section}
We proposed knowledge distillation via intermediate classifier heads (DIH). The intermediate classifiers convert the teacher's intermediate knowledge to the ``digestible'' output probabilities for the student. In DIH, the student---by distilling the knowledge from all the intermediate and teacher's classifier heads---has access to a heterogeneous source of knowledge to consume based on its representation capacity. 
We demonstrated the strengths and the generalizability of DIH by conducting extensive experiments on standard benchmarks and various student-teacher pairs of models.  We also investigated the reasons behind DIH's success by doing an extensive ablation study and deploying concepts such as information entropy, classifiers' confidence, and overthinking. Applying this framework in online knowledge distillation, when both student and teacher teach each other such as deep mutual learning \citep{zhang2018deep}, could be a promising direction for future studies.

\bibliographystyle{plain}
\bibliography{refbib}
\end{document}